\documentclass[letterpaper]{article} 
\usepackage{aaai2026}  
\usepackage{times}  
\usepackage{helvet}  
\usepackage{courier}  
\usepackage[hyphens]{url}  
\usepackage{graphicx} 
\usepackage{natbib}  
\usepackage{caption} 
\usepackage{algorithm}
\usepackage{algorithmic}

\usepackage{soul}
\usepackage{newfloat}
\usepackage{listings}

\usepackage{amsmath} 
\usepackage{amssymb}
\usepackage{tikz}
\usepackage{booktabs}

\usetikzlibrary{arrows.meta,positioning,shapes.misc,fit,calc}
\usepackage{pgfplots}
\pgfplotsset{compat=1.18}

\usetikzlibrary{patterns}

\DeclareCaptionStyle{ruled}{labelfont=normalfont,labelsep=colon,strut=off} 
\floatstyle{ruled}
\newfloat{listing}{tb}{lst}{}
\floatname{listing}{Listing}
\title{PEFT-DML: Parameter-Efficient Fine-Tuning Deep Metric Learning for Robust Multi-Modal 3D Object Detection in Autonomous Driving}
\author{
    Abdolazim Rezaei, 
    Mehdi Sookhak 
}
\affiliations{

    Department of Computer Science, Texas A\&M University, 
    Corpus Christi, TX 78412 USA \\
    arezaei@islander.tamucc.edu, m.sookhak@ieee.org
}

\usepackage{bibentry}

\begin{document}

\maketitle

\begin{abstract}
This study introduces PEFT-DML, a parameter-efficient deep metric learning framework for robust multi-modal 3D object detection in autonomous driving. Unlike conventional models that assume fixed sensor availability, PEFT-DML maps diverse modalities (LiDAR, radar, camera, IMU, GNSS) into a shared latent space, enabling reliable detection even under sensor dropout or unseen modality–class combinations. By integrating Low-Rank Adaptation (LoRA) and adapter layers, PEFT-DML achieves significant training efficiency while enhancing robustness to fast motion, weather variability, and domain shifts. Experiments on benchmarks nuScenes demonstrate superior accuracy.
\end{abstract}


\section{Introduction}

Reliable detection of moving 3D objects is fundamental for autonomous driving but remains challenged by fast motion, environmental variability, and sensor limitations. 
To overcome these issues, we propose PEFT-DML, which unifies LiDAR, radar, camera, IMU, and GNSS into a shared latent space. Using LoRA and adapters, PEFT-DML achieves robust, modality-agnostic detection with reduced training cost. 

The proposed PEFT-DML framework surpasses recent studies in multi-modal 3D object detection and cooperative perception. Unlike 3ML-DML framework \cite{dullerud2022integrated}, which requires fixed modality availability, PEFT-DML generalizes across unseen modality–class combinations through a unified latent space, enabling zero-shot cross-modal detection. In addition, CRKD \cite{zhao2024crkd} focuses on distillation between camera and radar but it is fragile when one sensor fails. PEFT-DML instead supports any subset of sensors which ensures robustness under partial sensor dropout. 


RoboFusion \cite{song2024robofusion} leverages computationally heavy Visual Foundation Models which limits efficiency. In contrast, PEFT-DML achieves comparable robustness via lightweight LoRA-based fine-tuning. 
The authors in \cite{chae2024towards} introduce weather-aware gating but still requires both modalities at inference whereas PEFT-DML performs robust detection even when one or more modalities are missing. 

\section{PEFT-DML Framework}

\begin{figure}[t]
\centering
\includegraphics[width=0.48\textwidth]{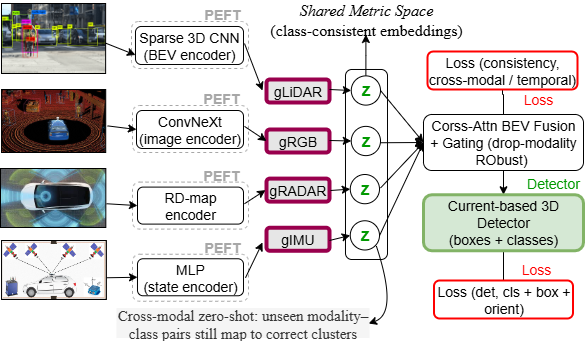} 

\caption{The PEFT-DML pipeline unifies various modalities into a shared latent space with LoRA and adapter layers.}

\label{fig:framework}
\end{figure}

\begin{table*}[ht]
\centering
\begin{tabular}{l c c c c c c c c}
\hline
Method &  mAP$\uparrow$ & NDS$\uparrow$ & mATE$\downarrow$ & mASE$\downarrow$ & mAOE$\downarrow$ & mAVE$\downarrow$ & mAAE$\downarrow$ \\
\hline
UVTR & 45.2 & 52.2 & 0.612 & 0.256 & 0.385 & 0.664 & 0.125 \\
X3KD & 45.6 & 56.1 & 0.506 & 0.253 & 0.414 & 0.366 & 0.131 \\
UniDistill & 29.6 & 39.3 & 0.637 & 0.257 & 0.492 & 1.084 & 0.167 \\
CRKD &  48.7 & 58.7 & 0.404 & 0.253 & 0.425 & 0.376 & 0.111 \\ 
RoboFusion &  54.3 & 67.1 & 0.338 & 0.229 & 0.382 & 0.367 & 0.102 \\
3D-LRF &  74.8 & 57.8 & 0.337 & 0.226 & 0.398 & 0.375 & 0.118 \\
PEFT-DML &  \textbf{62.2} & \textbf{71.7} & \textbf{0.316} & \textbf{0.206} & \textbf{0.346} & \textbf{0.339} & \textbf{0.093} \\
\hline
\end{tabular}
\caption{Table \ref{tab:metrics}: PEFT-DML achieves the best performance across all metrics,}
\label{tab:metrics}
\end{table*}



We propose \textbf{PEFT-DML}, a framework for robust 3D object detection in autonomous driving (Figure~\ref{fig:framework}). 
The model unifies heterogeneous modalities 
where backbone encoders remain frozen to preserve pretrained features, while lightweight \textit{LoRA} and adapter layers enable efficient fine-tuning. 
Each modality is mapped by a projection head into a normalized $d$-dimensional embedding, where intra-class features cluster closely and inter-class features remain distinct. 
Cross-attention and gating modules fuse embeddings, ensuring flexibility under sensor dropout. 
A detection head then predicts 3D bounding boxes and class labels.

Training is guided by a joint multi-objective loss 
$\mathcal{L} = \lambda_{\text{det}} \mathcal{L}_{\text{det}} + \lambda_{\text{met}} \mathcal{L}_{\text{metric}} + \lambda_{\text{cons}} \mathcal{L}_{\text{consistency}}$
where \textbf{Detection Loss} is $\mathcal{L}_{\text{det}} = \mathcal{L}_{\text{cls}} + \mathcal{L}_{\text{reg}} 
= \text{FocalCE}(y,\hat{y}) + \text{IoU}(b, \hat{b}) + \| o - \hat{o} \|_1$. This combines focal classification loss, IoU-based bounding box regression, and orientation regression.

\textbf{Metric Alignment Loss}, is 
$\mathcal{L}_{\text{metric}} = \max \big(0, d(z_i, z_j) - d(z_i, z_k) + \alpha \big)$. 
A triplet loss encourages embeddings $z_i$ and $z_j$ from the same class to be closer than $z_i$ and $z_k$ from different classes.

In addition, \textbf{Consistency Loss} is 
$\mathcal{L}_{\text{consistency}} = \| z_t - z_{t+1} \|_2^2$ which enforces temporal stability across adjacent frames and consistency across modalities.

\noindent Together, these objectives enable \textbf{cross-modal zero-shot generalization} by mapping them into the shared latent space and comparing them with embeddings from known modalities. 

\section{Experiments}
Experiments on nuScenes dataset demonstrate that PEFT-DML achieves superior accuracy, robustness, and parameter efficiency compared to state-of-the-art baselines.

\begin{figure}[t]
\centering
\begin{tikzpicture}
\begin{axis}[
    ybar,
    bar width=9pt,
    width=1\linewidth, height=6cm,
    ymin=0, ymax=70,
    enlarge x limits=0.2,
    ylabel={AP3D, IoU=0.5},
    symbolic x coords={Total,Normal,Fog,Rain,Snow},
    xtick=data,
    x tick label style={rotate=45,anchor=east},
    legend style={at={(0.5,1.05)}, anchor=south, legend columns=-1}
]

\addplot+[ybar, bar shift=-9pt, pattern=horizontal lines, pattern color=black] coordinates {
    (Total,14.1) (Normal,19.7) (Fog,15.9) (Rain,13.0) (Snow,21.0)
};
\addplot+[ybar, bar shift=0pt, pattern=north east lines, pattern color=black] coordinates {
    (Total,37.8) (Normal,39.8) (Fog,59.8) (Rain,28.2) (Snow,50.7)
};
\addplot+[ybar, bar shift=9pt, pattern=dots, pattern color=black] coordinates {
    (Total,49.0) (Normal,50.2) (Fog,62.0) (Rain,35.0) (Snow,58.0)
};

\legend{RoboFusion, 3D-LRF, \textbf{PEFT-DML (Ours)}}
\end{axis}
\end{tikzpicture}
\caption{Comparison over different climate conditions.}
\label{fig:falls}
\end{figure}
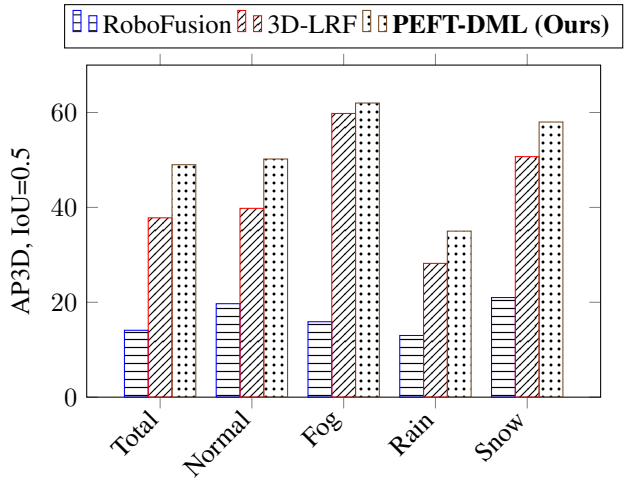



\begin{figure}[t]
\centering
\begin{tikzpicture}
\begin{axis}[
    width=0.9\linewidth,height=5.5cm,
    xmode=log, log basis x=10, 
    xlabel={Trainable Params (Millions, log)}, ylabel={Metric $M$},
    xmin=3, xmax=200, ymin=62, ymax=73, grid=both,
    legend pos=south east
]
\addplot+[mark=*] coordinates {(113,66.0) (130,69.8)}; \addlegendentry{Full FT}
\addplot+[mark=square*] coordinates {(3.5,66.5) (7.0,69.8) (14.0,71.2)}; \addlegendentry{PEFT-DML ($r{=}\{4,8,16\}$)}
\end{axis}
\end{tikzpicture}
\caption{PEFT-DML achieves nearly the same or slightly higher accuracy than Full Fine-Tuning (Full-FT) while updating less than 10\% of the parameters. 
}
\label{fig:accuracy}
\end{figure}
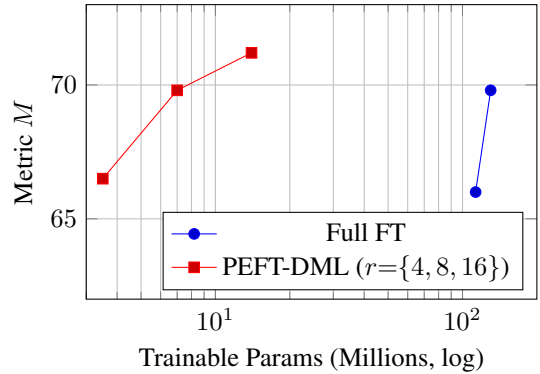

Figure \ref{fig:falls}: PEFT-DML consistently achieves the highest AP3D scores across all weather conditions, demonstrating superior robustness compared to RoboFusion and 3D-LRF. Furthermore, PEFT-DML in Figure \ref{fig:accuracy} achieves higher accuracy than full fine-tuning while requiring far fewer trainable parameters, demonstrating superior parameter efficiency.
Table \ref{tab:metrics} compares PEFT-DML with six recent methods across the nuScenes benchmark using detection and error metrics. The results demonstrate that PEFT-DML consistently outperforms all baselines. It achieves the highest mAP (62.2) and NDS (71.7), reflecting superior detection accuracy and overall performance. In terms of localization and geometry, PEFT-DML attains the lowest mATE (0.316) and mASE (0.206), indicating more precise and stable bounding boxes. It also outperforms others in orientation and velocity estimation, with the lowest mAOE (0.346), mAVE (0.339), and mAAE (0.993). 

\section{Conclusion}
In conclusion, the proposed PEFT-DML framework delivers a modality-agnostic 3D object detection for autonomous driving. By unifying heterogeneous sensors in a shared latent space and employing parameter-efficient fine-tuning, it outperforms other models.

\bibliography{References}

@inproceedings{chae2024towards,
  title={Towards robust 3d object detection with lidar and 4d radar fusion in various weather conditions},
  author={Chae, Yujeong and Kim, Hyeonseong and Yoon, Kuk-Jin},
  booktitle={Proceedings of the IEEE/CVF Conference on Computer Vision and Pattern Recognition},
  pages={15162--15172},
  year={2024}
}

@article{song2024robofusion,
  title={RoboFusion: Towards robust multi-modal 3D object detection via SAM},
  author={Song, Ziying and Zhang, Guoxing and Liu, Lin and Yang, Lei and Xu, Shaoqing and Jia, Caiyan and Jia, Feiyang and Wang, Li},
  journal={arXiv preprint arXiv:2401.03907},
  year={2024}
}

@inproceedings{zhao2024crkd,
  title={Crkd: Enhanced camera-radar object detection with cross-modality knowledge distillation},
  author={Zhao, Lingjun and Song, Jingyu and Skinner, Katherine A},
  booktitle={Proceedings of the IEEE/CVF Conference on Computer Vision and Pattern Recognition},
  pages={15470--15480},
  year={2024}
}

@inproceedings{dullerud2022integrated,
  title     = {An Integrated Multi-Label Multi-Modal Framework in Deep Metric Learning},
  author    = {Natalie Dullerud and Karsten Roth and Haoran Zhang and Qixuan Jin and Thomas Hartvigsen and Marzyeh Ghassemi},
  booktitle = {Proceedings of the International Conference on Learning Representations (ICLR)},
  year      = {2022},
  note      = {Withdrawn submission}
}

\end{document}